\pdfoutput=1

\documentclass[10pt,twocolumn,letterpaper]{article}

\usepackage{cvpr}
\usepackage{times}
\usepackage{epsfig}
\usepackage{graphicx}
\usepackage{amsmath}
\usepackage{amssymb}

\usepackage{multirow}

\cvprfinalcopy % *** Uncomment this line for the final submission

\def\cvprPaperID{430} % *** Enter the CVPR Paper ID here

\ifcvprfinal\fi
\begin{document}

%%%%%%%%% TITLE
\title{GeoNet: Deep Geodesic Networks for Point Cloud Analysis}

\author{Tong He$^{1}$,\ \ Haibin Huang$^{2}$,\ \ Li Yi$^{3}$,\ \ Yuqian Zhou$^{4}$,\ \ Chihao Wu$^{2}$,\ \ Jue Wang$^{2}$,\ \ Stefano Soatto$^{1}$\\
% UCLA Vision Lab\\
$^{1}$UCLA\ \ \ \ $^{2}$Megvii (Face++)\ \ \ \ $^{3}$Stanford\ \ \ \ $^{4}$UIUC\\
}

% \author{Tong He^{1}\\
% \and
% Haibin Huang^{2}\\
% \and
% Li Yi^{3}\\
% \and
% Yuqian Zhou^{4}\\
% \and
% Chihao Wu^{2}\\
% \and
% Jue Wang^{2}\\
% \and
% Stefano Soatto^{1}\\

\maketitle
%\thispagestyle{empty}

%%%%%%%%% ABSTRACT
\begin{abstract}
\vspace{-2mm}

Surface-based geodesic topology provides strong cues for object semantic analysis and geometric modeling. However, such connectivity information is lost in point clouds. Thus we introduce GeoNet, the first deep learning architecture trained to model the intrinsic structure of surfaces represented as point clouds. To demonstrate the applicability of learned geodesic-aware representations, we propose fusion schemes which use GeoNet in conjunction with other baseline or backbone networks, such as PU-Net and PointNet++, for down-stream point cloud analysis. Our method improves the state-of-the-art on multiple representative tasks that can benefit from understandings of the underlying surface topology, including point upsampling, normal estimation, mesh reconstruction and non-rigid shape classification.

%We demonstrate qualitative and quantitative evaluations on several benchmarks where our method surpasses the state-of-the-art works.

%our method surpasses the state-of-the-art on multiple representative tasks that can benefit from understandings of the underlying surface topology, including point upsampling, normal estimation, mesh reconstruction and non-rigid shape classification.

 %Geodesic distance based point cloud topology provides important cues for both point cloud semantic analysis and geometric modeling. However, 3D points in a point could are represented independently in Euclidean coordinates. To model the underlying point connectivity, we propose the first data-driven method, geodesic neighborhood estimation network (GeoNet), for point cloud topology recovery by supervised geodesic training. GeoNet can recover various topological patterns of a point cloud and surpasses traditional KNN-Graph based shortest path methods. To further leverage the learned point relationship for down-stream tasks, we present two novel geodesic fusion frameworks which integrate GeoNet into widely used deep networks, such as PointNet++, for point set feature learning. Our method is benchmarked on multiple datasets and outperforms the state-of-the-art methods on representative tasks that require intrinsic understandings of the underlying point cloud topology, including point upsampling, normal estimation, mesh reconstruction and non-rigid shape classification.
 \vspace{-3mm}
\end{abstract}
%%%%%%%%% INTRODUCTION
\section{Introduction}

\begin{figure}%[htb]
\centering
\resizebox{1.0 \columnwidth}{!}{
% \makebox[1 \textwidth][c]{

% \includegraphics[height=6.5cm]{pr_curves.pdf}
\includegraphics[height=6.5cm]{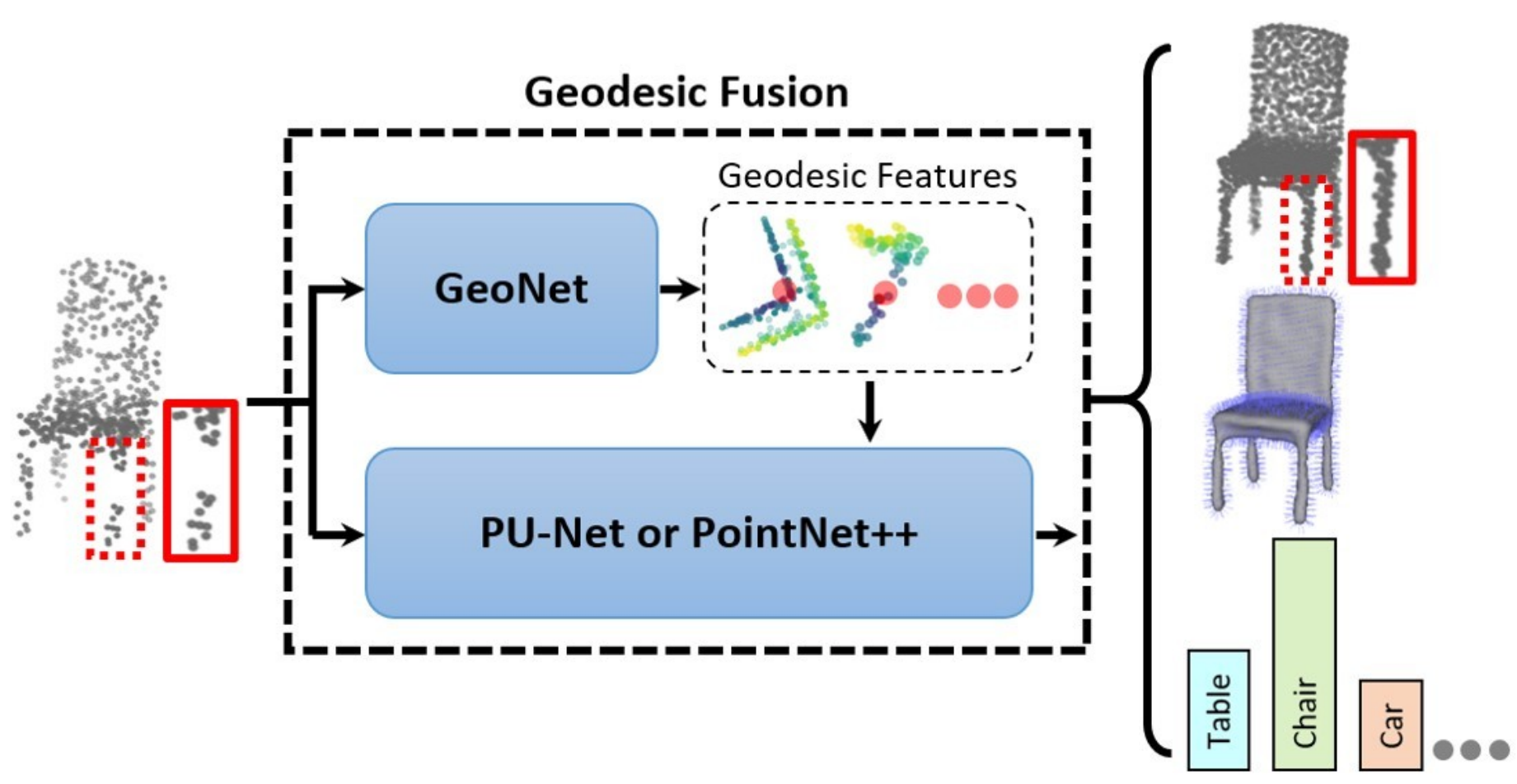}

}
\caption{Our method takes a point cloud as input, and outputs representations used for multiple tasks including upsampling, normal estimation, mesh reconstruction, and shape classification.}
\label{teaser}
% \vspace{-4mm}
\end{figure}

\noindent Determining neighborhood relationship among points in a point cloud, known as topology estimation, is an important problem since it indicates the underlying point cloud structure, which could further reveal the point cloud semantics and functionality. Consider the red inset in Fig.~\ref{teaser}: the two clusters of points, though seemingly disconnected, should indeed be connected to form a chair leg, which supports the whole chair. On the other hand, the points on opposite sides of a chair seat, though spatially very close to each other, should not be connected to avoid confusing the sittable upper surface with the unsittable lower side. Determining such topology appears to be a very low-level endeavor but in reality it requires global, high-level knowledge, making it a very challenging task. Still, from the red inset in Fig.~\ref{teaser},  we could draw the conclusion that the two stumps are connected only after we learn statistical regularities from a large number of point clouds and observe many objects of this type with connected elongated vertical elements extending from the body to the ground. This motivates us to adopt a learning approach to capture the topological structure within point clouds.

Our primary goals in this paper are to develop representations of point cloud data that are informed by the underlying surface topology as well as object geometry, and propose methods that leverage the learned topological features for geodesic-aware point cloud analysis. The representation should capture various topological patterns of a point cloud and the method of leveraging these geodesic features should not alter the data stream, so our representation can be learned jointly and used in conjunction with the state-of-the-art baseline or backbone models (e.g. PU-Net, PointNet++~\cite{yu2018pu,qi2017pointnet,qi2017pointnet++}) that feed the raw data through, with no information loss to further stages of processing.

%%%%%%%%%%%%%%%%%%%%%%%%%%%%%%%%%%%%%%%%%%%%%%%%%%%%%%%%%%%%%%%%%%%%%%%%%%%%%%%%%%%%%%
%%%%%%%%%%%%%%%%%%%%%%%%%%%%%%%%%%%%%%%%%%%%%%%%%%%%%%%%%%%%%%%%%%%%%%%%%%%%%%%%%%%%%%

\begin{figure*}
\vspace{-1mm}
\centering
\resizebox{1.0 \textwidth}{!}{
% \makebox[1 \textwidth][c]{

% \includegraphics[height=6.5cm]{pr_curves.pdf}
\includegraphics[height=6.5cm]{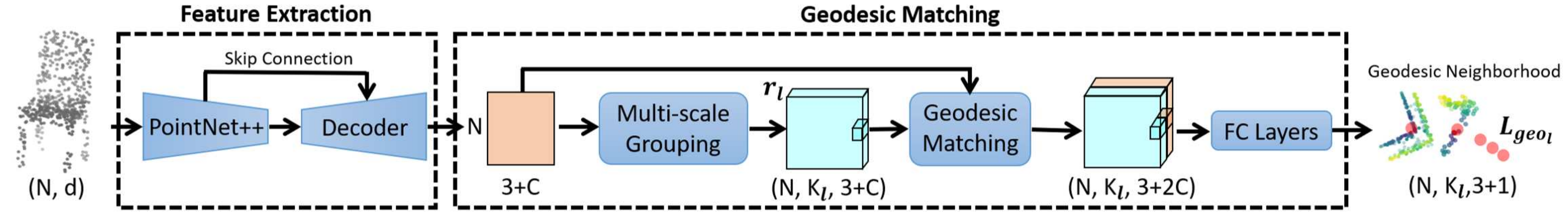}

}
\caption{GeoNet: geodesic neighborhood estimation network.}
\label{top_net_pipeline}
% \vspace{-3mm}
\end{figure*}

For the first goal, we propose a geodesic neighborhood estimation network (GeoNet) to learn deep geodesic representations using the ground truth geodesic distance as supervision signals. As illustrated in Fig.~\ref{top_net_pipeline}, GeoNet consists of two modules: an autoencoder that extracts a feature vector for each point and a geodesic matching (GM) layer that acts as a learned kernel function for estimating geodesic neighborhoods using the latent features. Due to the supervised geodesic training process, intermediates features of the GM layer contain rich information of the point cloud topology and intrinsic surface attributes. We note that the representation, while trained on geodesic distances, does not by construction produce geodesics (e.g. symmetry, triangle inequality, etc.). The goal of the representation is to inform subsequent stages of processing of the global geometry and topology, and is not to conduct metric computations directly.

\begin{figure*}
\vspace{-2mm}
\centering
\resizebox{1.0 \textwidth}{!}{
% \makebox[1 \textwidth][c]{

% \includegraphics[height=6.5cm]{pr_curves.pdf}
\includegraphics[height=6.5cm]{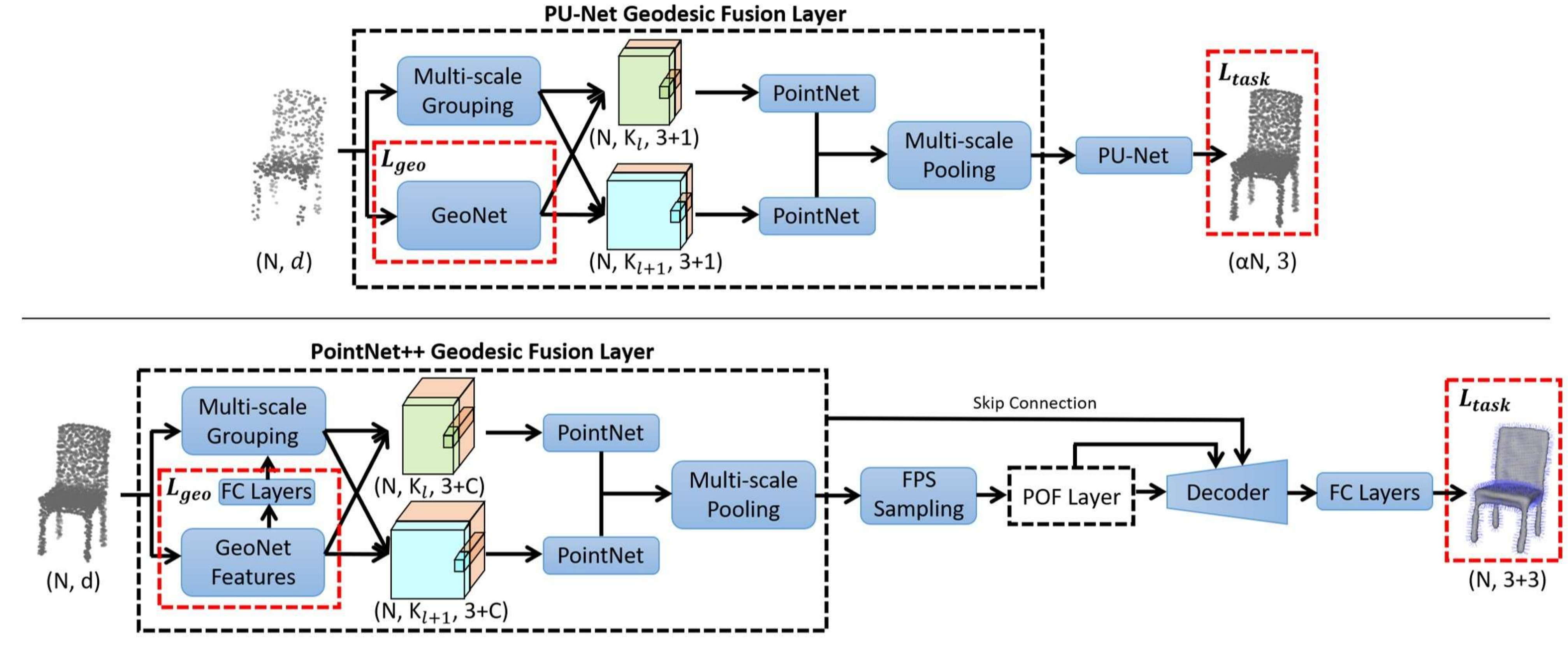}

}
\caption{PU-Net (top) and PointNet++ (bottom) geodesic fusion  architectures.}
\label{geo_fusion_net}
% \vspace{-2mm}
\end{figure*}

For the second task, as shown in Fig.~\ref{geo_fusion_net}, we propose geodesic fusion schemes to integrate GeoNet into the state-of-the-art network architectures designed for different tasks. Specifically, we present PU-Net fusion (PUF) for point cloud upsampling, and PointNet++ fusion (POF) for normal estimation, mesh reconstruction as well as non-rigid shape classification. Through experiments, we demonstrate that the learned geodesic representations from GeoNet are beneficial for both geometric and semantic point cloud analyses.

In summary, in this paper we propose an approach for learning deep geodesic-aware representations from point clouds and leverage the results for various point set analyses. Our contributions are:
\begin{itemize}
\item We present, to the best of our knowledge, the first deep learning method, GeoNet, that ingests point clouds and learns representations which are informed by the intrinsic structure of the underlying point set surfaces.
\item To demonstrate the applicability of learned geodesic representations, we develop network fusion architectures that incorporate GeoNet with baseline or backbone networks for geodesic-aware point set analysis.
% \item We conduct extensive experiments as well as analyses to test our geodesic fusion methods on both geometric and semantic point set tasks using standard datasets and outperform the state-of-the-art methods.
\item Our geodesic fusion methods are benchmarked on multiple geometric and semantic point set tasks using standard datasets and outperform the state-of-the-art methods.
% \vspace{-1mm}
\end{itemize}

%One could argue that point clouds often come with photometric signatures, e.g. Lidar or radar reflectance, or RGB values in a structured light camera, and these can be informative of scene topology. This can be done, but it is not our focus here. Instead, we focus on point clouds alone as photometric signatures could be absent ({\em e.g.}, in point clouds derived from CAD models) or dependent on the unknown illuminant. \tong{I prefer not to mention this bullet, which might confuse the reviewers. Though combining images with point clouds is more informative then just studying the point cloud alone, the combined analysis is not what people commonly do for point cloud related tasks. Thus mentioning this bullet will probably make people think that our paper has some "story" to say about it.}

%{While topology is a low-level cue, it both influences and requires high-level understanding of objects. We do not alter the data stream, so our representation can be used in conjunction to skip connections that feed the raw data through, with no information loss to further stages of processing.} \tong{This point is not quite clear to me. I think this bullet is more suitable to be mentioned while introducing the two geodesic fusion methods.}

%%%%%%%%% RELATED WORK
%%%%%%%%%%%%%%%%%%%%%%%%%%%%%%%%%%%%%%%%%%%%%%%%%%%%%%%%%%%%%%%%%%%%%%%%%%%%%%%%%%%%%%
%%%%%%%%%%%%%%%%%%%%%%%%%%%%%%%%%%%%%%%%%%%%%%%%%%%%%%%%%%%%%%%%%%%%%%%%%%%%%%%%%%%%%%

\section{Related work}
We mainly review traditional graph-based methods for geodesic distance computation, as well as general works on point cloud upsampling, normal estimation, and non-rigid shape classification, as we are unaware of other prior works on point cloud-based deep geodesic representation learning.

\textbf{Geodesic distance computation.} There are two types of methods: some allow the path to traverse mesh faces~\cite{sharir1986shortest,mitchell1987discrete,chen1990shortest,garland1997surface,surazhsky2005fast,xin2009improving,crane2013geodesics} for accurate geodesic distance computation, while others find approximate solutions via shortest path algorithms constrained on graph edges~\cite{Dijkstra,floyd1962algorithm,johnson1977efficient}. For the first type, an early method~\cite{sharir1986shortest} suggests a polynomial algorithm of time $O(n^3logn)$ where $n$ is the number of edges, but their method is restricted to a convex polytope. Based on Dijkstra's algorithm~\cite{Dijkstra}, \cite{mitchell1987discrete} improves the time complexity to $O(n^2logn)$ and extends the method to an arbitrary polyhedral surface. Later, \cite{chen1990shortest} proposes an $O(n^2)$ approach using a set of windows on the polyhedron edges to encode the structure of the shortest path set. By filtering out useless windows, \cite{xin2009improving} further speeds up the algorithm. Then~\cite{crane2013geodesics} introduces a heat method via solving a pair of standard linear elliptic problems. As for graph edge-based methods, typical solutions include Dijkstra's~\cite{Dijkstra}, Floyd-Warshall~\cite{floyd1962algorithm} and Johnson's algorithms~\cite{johnson1977efficient}, which have much lower time complexity than the surface traversing methods. For a 20000-vertex mesh, computing its all-pair geodesic distances can take several days using~\cite{xin2009improving} while~\cite{johnson1977efficient} only uses about 1 minute on CPU. When a mesh is dense, the edge-constrained shortest path methods generate low-error geodesic estimates. Thus in our work, we apply~\cite{johnson1977efficient} to compute the ground truth geodesic distance. % It is important to notice that methods described above all require a given mesh, which is not available from widely used sensors for 3D data collection (e.g. depth camera, Lidar, etc.), and this motivates our research.

\textbf{Point upsampling.} Previous methods can be summarized into two categories. i) Optimization based methods~\cite{alexa2003computing,lipman2007parameterization,huang2009consolidation}, championed by~\cite{alexa2003computing}, which interpolates a dense point set from vertices of a Voronoi diagram in the local tangent space. Then~\cite{lipman2007parameterization} proposes a locally optimal projection (LOP) operator for point cloud resampling and mesh reconstruction leveraging an $L_1$ median. For improving robustness to point cloud density variations, \cite{huang2009consolidation} presents a weighted LOP. These methods all make strong assumptions, such as surface smoothness, and are not data-driven, and therefore have limited applications in practice. ii) Deep learning based methods. To apply the (graph) convolution operation, many of those methods first voxelize a point cloud into regular volumetric grids~\cite{wu20153d,wu2016learning,han2017high,dai2017shape} or instead use a mesh~\cite{defferrard2016convolutional,yi2017syncspeccnn}. While voxelization introduces discretization artifacts and generates low resolution voxels for computational efficiency, mesh data can not be trivially reconstructed from a sparse and noisy point cloud. To directly upsample a point cloud, PU-Net~\cite{yu2018pu} learns multilevel features for each point and expands the point set via a multibranch convolution unit implicitly in feature space. But PU-Net is based on Euclidean space and thus does not leverage the underlying point cloud surface attributes in geodesic space, which we show in this paper are important for upsampling.

\textbf{Normal estimation.} A widely used method for point cloud normal estimation is to analyze the variance in a tangential plane of a point and find the minimal variance direction by Principal Component Analysis (PCA)~\cite{hoppe1992surface,jolliffe2011principal}. But this method is sensitive to the choice of the neighborhood size, namely, large regions can cause over-smoothed surfaces and small ones are sensitive to noises. To improve robustness, methods based on fitting higher-order shapes have been proposed~\cite{guennebaud2007algebraic,cazals2005estimating,amenta1999surface}. %, such as sphere~\cite{guennebaud2007algebraic}, quadratic surface~\cite{cazals2005estimating}, Delaunay triangulation~\cite{amenta1999surface}.
However, these methods require careful parameter tuning at the inference time and only estimate normal orientation up to sign. %For consistent orientation estimation, \cite{hoppe1992surface,xie2003piecewise,liu2014mendable} propagate the normal orientation of neighboring points in a greedy manner. Others methods that can determine the normal orientation include~\cite{zhao2001fast,giraudot2013noise}, which leverage volumetric 3D representations such as a signed distance function. As explained in point upsampling, processing voxel grids is computational heavy and has discretization artifacts.
Thus, so far robust estimation for oriented normal vectors using traditional methods is still challenging, especially across different noise levels and shape structures. There are only few data-driven methods that are able to integrate normal estimation and orientation alignment into a unified pipeline~\cite{guerrero2018pcpnet,qi2017pointnet++}. They take a point cloud as input and directly regress oriented normal vectors, but these methods are not designed to learn geodesic topology-based representations that capture the intrinsic surface features for better normal estimation.

\textbf{Non-rigid shape classification.} Classifying the point cloud of non-rigid objects often consists of two steps: extracting intrinsic features in geodesic space and applying a classifier (e.g. SVM, MLP, etc.). Some commonly used features include wave kernel signatures~\cite{aubry2011wave}, heat kernel signatures~\cite{sun2009concise}, spectral graph wavelet signatures~\cite{masoumi2016spectral}, Shape-DNA~\cite{reuter2006laplace}, etc. For example~\cite{luciano2018deep} uses geodesic moments and stacked sparse autoencoders to classify non-rigid shapes, such as cat, horse, spider, etc. The geodesic moments are feature vectors derived from the integral of the geodesic distance on a shape, while stacked sparse autoencoders are deep neural networks consisting of multiple layers of sparse autoencoders. However, the above methods all require knowing graph-based data, which is not available from widely used sensors (e.g. depth camera, Lidar, etc.) for 3D data acquisition. Though PointNet++~\cite{qi2017pointnet++} is able to directly ingest a point cloud and conduct classification, it is not designed to model the geodesic topology of non-rigid shapes and thus its performance is inferior to traditional two-step methods which heavily reply on the offline computed intrinsic surface features.
%%%%%%%%% PROBLEM STATEMENT
%%%%%%%%%%%%%%%%%%%%%%%%%%%%%%%%%%%%%%%%%%%%%%%%%%%%%%%%%%%%%%%%%%%%%%%%%%%%%%%%%%%%%%
%%%%%%%%%%%%%%%%%%%%%%%%%%%%%%%%%%%%%%%%%%%%%%%%%%%%%%%%%%%%%%%%%%%%%%%%%%%%%%%%%%%%%%

%%%%%%%%% METHOD
\vspace{-3mm}
\section{Method}

\subsection{Problem Statement}
$\chi =\{x_{i}\}$ denotes a point set with $x_i \in {\mathbb R}^d$ and $i = 1, \dots, N$. Although the problem and the method developed are general, we focus on the case $d = 3$ using only Euclidean coordinates as input. A neighborhood subset within radius $r$ from a point $x_i$ is denoted $B_r(x_i) = \{x_j | d_E(x_i,x_j)\leq r\}$ where $d_E(x_i,x_j) \in {\mathbb R}$ is the Euclidean (embedding) distance between $x_i$ and $x_j$. The cardinality of $B_r(x_i)$ is $K$. The corresponding geodesic distance set around $x_{i}$ is called $G_r(x_i) = \{g_{ij}=d_G(x_i,x_j) | x_j \in B_r(x_i)\}$ where $d_G \in {\mathbb R}$ means the geodesic distance. Our goal is to learn a function $f: x_i \mapsto G_r(x_i)$ that maps each point to (an approximation of) the geodesic distance set $G_r(x_i)$ around it.

\subsection{Method}
We introduce GeoNet, a network trained to learn the function $f$ defined above. It consists of an autoencoder with skip connections, followed by a multi-scale Geodesic Matching (GM) layer, leveraging latent space features $\{\psi (x_i)\} \subseteq {\mathbb R^{3+C}}$ of the point set. GeoNet is trained in a supervised manner using ground truth geodesic distances between points in the set $\chi$. To demonstrate the applicability of learned deep geodesic-aware representations from GeoNet, we test our approach on typical tasks that require understandings of the underlying surface topology, including point cloud upsampling, surface normal estimation, mesh reconstruction, and non-rigid shape classification. To this end, we leverage the existing state-of-the-art network architectures designed for the aforementioned problems. Specifically, we choose PU-Net as the baseline network for point upsampling and PointNet++ for other tasks. The proposed geodesic fusion methods, called PU-Net fusion (PUF) and PointNet++ fusion (POF), integrate GeoNet with the baseline or backbone models to conduct geodesic-aware point set analysis.
% \vspace{-4mm}

\subsection{Geodesic Neighborhood Estimation}
As illustrated in Fig.~\ref{top_net_pipeline}, GeoNet consists of two modules: an autoencoder that extracts a feature vector $\psi (x_i)$ for each point $x_i \in \chi$ and a GM layer that acts as a learned geodesic kernel function for estimating $G_r(x_i)$ using the latent features.

\textbf{Feature Extraction.} We use a variant of PointNet++, which is a point set based hierarchical and multi-scale function, for feature extraction. It maps an input point set $\chi$ to a feature set $\{\varphi (x_i) | x_i \in \widetilde{\chi} \}$ where $ \varphi (x_i) \in {\mathbb R^{3+\widetilde{C}}}$ is a concatenation of the $xyz$ coordinates and the $\widetilde{C}$ dimensional embedding of $x_i$, and $\widetilde{\chi}$ is a sampled subset of $\chi$ by farthest-point sampling. To recover features $\{\psi (x_i)\}$ for the point cloud $\chi$, we use a decoder with skip connections. The decoder consists of recursively applied tri-linear feature interpolators, shared fully connected (FC) layers, ReLU and Batch Normalization. The resulting $(N, 3+C)$ tensor is then fed into the GM layer for geodesic neighborhood estimation.

\textbf{Geodesic Matching.} We group the latent features $\psi (x_i)$ into neighborhood feature sets $F_{r_l}(x_i)=\{\psi(x_j) | x_j \in B_{r_l}(x_i) \}$, under multiple radius scales $r_l$. At each scale $r_l$ we set a maximum number of neighborhood points $K_l$, and thus produce a tensor of dimension $(N, K_l, 3+C)$. The grouped features, together with the latent features, are sent to a geodesic matching module, where $\psi (x_i)$ is concatenated with $\psi (x_j)$ for every $x_j \in B_{r_l}(x_i)$. The resulting feature $\xi_{ij} \in {\mathbb R^{3+2C}}$ becomes the input to a set of shared FC layers with ReLU, Batch Normalization and Dropout. As demonstrated in~\cite{han2015matchnet}, the multilayer perceptron (MLP) acts as a kernel function that maps $\xi_{ij}$ to an approximation of the geodesic distance, $\hat{g}_{ij}$. Finally, the GM layer yields $G_{r_l}(x_i)$ for each point of the input point cloud $\chi$. We use a multi-scale $L_1$ loss $L_{geo}=\sum_{l}L_{geo_l}$ to compare the ground truth geodesic distances to their estimates:

\begin{align} \label{loss_geo}
\begin{split}
L_{geo_l} = \sum_{x_i \in \chi}\sum_{x_j \in B_{r_l}(x_i)}\frac{\left | g_{ij} - \hat{g}(x_i,x_j) \right |}{NK_{l}}
\end{split}
\end{align}

%%%%%%%%%%%%%%%%%%%%%%%%%%%%%%%%%%%%%%%%%%%%%%%%%%%%%%%%%%%%%%%%%%%%%%%%%%%%%%%%%%%%%%
\begin{figure*}
\vspace{-3mm}
\centering
\resizebox{1 \textwidth}{!}{
% \makebox[1 \textwidth][c]{

% \includegraphics[height=6.5cm]{pr_curves.pdf}
\includegraphics[height=6.5cm]{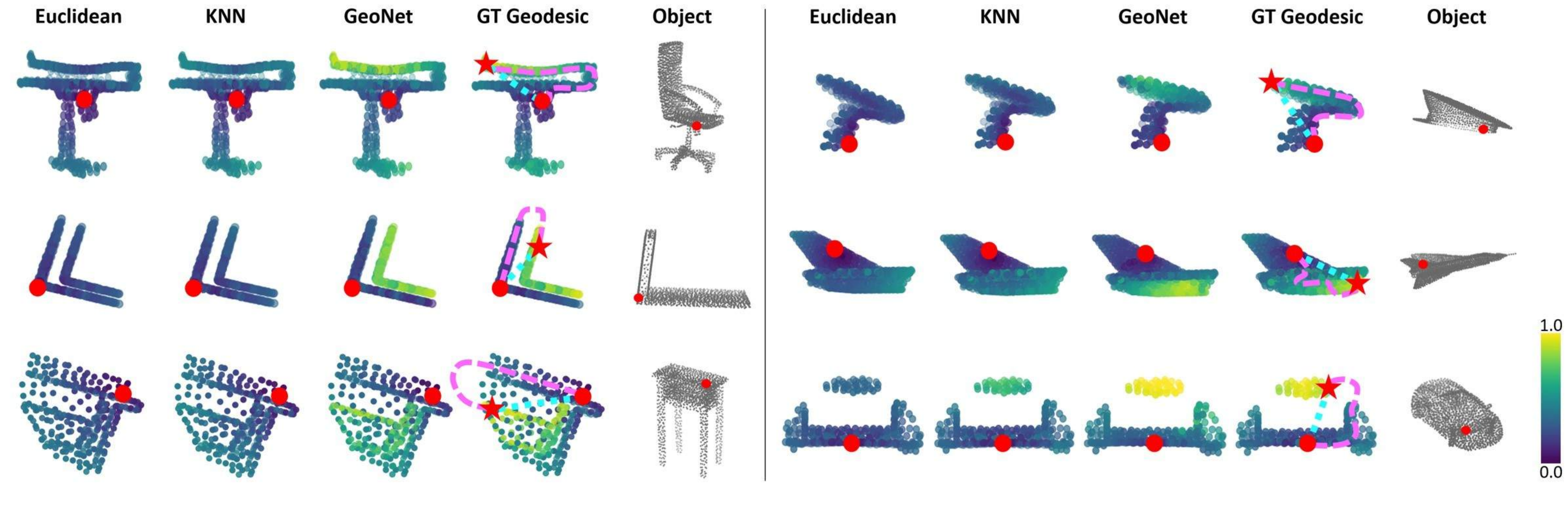}

}
\caption{Representative results of geodesic neighborhood estimation. Red dots indicate the reference point and stars represent target points selected for the purpose of illustration. Points in dark-purple are closer to the reference point than those in bright-yellow. Shortest paths between the reference point and the target point in euclidean space are colored in sky-blue. Topology-based geodesic paths are in pink.}
\label{geo_est_shape_net}
% \vspace{-2mm}
\end{figure*}
%%%%%%%%%%%%%%%%%%%%%%%%%%%%%%%%%%%%%%%%%%%%%%%%%%%%%%%%%%%%%%%%%%%%%%%%%%%%%%%%%%%%%%

\subsection{Geodesic Fusion}

To demonstrate how the learned geodesic representations can be used for point set analysis, we propose fusion methods based on the state-of-the-art (SOTA) network architectures for different tasks. For example, PU-Net is the SOTA upsampling method and thus we propose PUF that uses PU-Net as the baseline network to conduct geodesic fusion for point cloud upsampling. With connectivity information provided by the estimated geodesic neighborhoods, our geodesic-fused upsampling network can better recover topological details, such as curves and sharp structures, than PU-Net. We also present POF leveraging PointNet++ as the fusion backbone, and demonstrate its effectiveness on both geometric and semantic tasks where PointNet++ shows the state-of-the-art performance. % that can benefit from understandings of the topological surface attributes, such as normal estimation, mesh reconstruction and non-rigid shape classification.

% To demonstrate how the learned geodesic representations can be used for point cloud analysis, we propose two fusion methods, one shallow, one deep, called SGF and DGF respectively. We first propose SGF and test it on point cloud upsampling. With connectivity information provided by the geodesic neighborhoods, our geodesic-fused upsampling network can better recover the underlying point set topology than the state-of-the-art method PU-Net. We also present DGF, which tightly integrates GeoNet into backbone networks (e.g. PointNet, PointNet++) for both geometric and semantic tasks that require understandings of the underlying surface topology: normal estimation, mesh reconstruction and non-rigid shape classification.

\textbf{PU-Net Geodesic Fusion.} A PUF layer, as illustrated in Fig.~\ref{geo_fusion_net} (top), takes a $(N, d)$ point set as input and sends it into two branches: one is a multi-scale Euclidean grouping layer, and the other is GeoNet. At each neighborhood scale $r_l$, the grouped point set $B_{r_l}(x_i)$ is fused with the estimated geodesic neighborhood $G_{r_l}(x_i)$ to yield $S_{r_l}(x_i) = \{(x_j,g_{ij})|x_j \in B_{r_l}(x_i)\}$ with $(x_j,g_{ij}) \in {\mathbb R^{d+1}}$. Then the $(N, K_l, d+1)$ fused tensor is fed to a PointNet to generate a $(N, C_l)$ feature tensor which will be stacked with features from other neighborhood scales. The remaining layers are from PU-Net. As indicated by the red rectangles in Fig.~\ref{geo_fusion_net}, the total loss has two weighted terms:

\begin{align} \label{loss_geo_fusion}
\begin{split}
L=L_{geo}+\lambda L_{task}
\end{split}
\end{align}

\noindent where $L_{geo}$ is for GeoNet training~\eqref{loss_geo}, $\lambda$ is a weight and $L_{task}$, in general, is the loss for the current task that we are targeting. In this case, the goal is point cloud upsampling: $L_{task}=L_{up}(\theta)$ where $\theta$ indicates network parameters. PUF upsampling takes a randomly distributed sparse point set $\chi$ as input and generates a uniformly distributed dense point cloud $\hat{P} \subseteq {\mathbb R^{3}}$. The upsampling factor is $\alpha = \frac{\left | P \right |}{\left | \chi \right |}$:

\begin{align} \label{loss_upsampling}
\begin{split}
L_{up}(\theta) = L_{EMD}(P,\hat{P}) + \lambda_1 L_{rep}(\hat{P}) + \lambda_2 \left \| \theta  \right \|^{2}
\end{split}
\end{align}

\noindent in which the first term is the Earth Mover Distance (EMD) between the upsampled point set $\hat{P}$ and the ground truth dense point cloud $P$:

\begin{align} \label{loss_emd}
\begin{split}
L_{EMD}(P,\hat{P})=\underset{\phi:\hat{P}\rightarrow P}{min}\sum_{p_i \in \hat{P}}\left \| p_i-\phi(p_i) \right \|^{2}
\end{split}
\end{align}

\noindent where $\phi:\hat{P}\rightarrow P$ indicates a bijection mapping.

The second term in~\eqref{loss_upsampling} is a repulsion loss which promotes a uniform spatial distribution for $\hat{P}$ by penalizing close point pairs:

\begin{align} \label{loss_rep}
\begin{split}
L_{rep}(\hat{P}) = \sum_{p_i \in \hat{P}}\sum_{p_j \in \widetilde{P}_i}\eta (\left \| p_i - p_j \right \|) \omega (\left \| p_i - p_j \right \|)
\end{split}
\end{align}

\noindent where $\widetilde{P}_i$ is a set of \textit{k}-nearest neighbors of $p_i$, $\eta(r)=-r$ penalizes close pairs $(p_i, p_j)$, and $\omega(r) = e^{-r^2/h^2}$ is a fast-decaying weight function with some constant $h$~\cite{huang2009consolidation,lipman2007parameterization}.

%%%%%%%%%%%%%%%%%%%%%%%%%%%%%%%%%%%%%%%%%%%%%%%%%%%%%%%%%%%%%%%%%%%%%%%%%%%%%%%%%%%%%%
\begin{table}
\centering
\resizebox{0.95 \columnwidth}{!}{
% \makebox[\columnwidth][c]{

\begin{tabular}{lc|r|r|r|r|c|}
\cline{3-7}
                                                    & \multicolumn{1}{r|}{} & \multicolumn{1}{c|}{\textit{K-3}} & \multicolumn{1}{c|}{\textit{K-6}} & \multicolumn{1}{c|}{\textit{K-12}} & \multicolumn{1}{c|}{\textit{Euc}} & \multicolumn{1}{c|}{GeoNet} \\ \hline
\multicolumn{1}{|l|}{\multirow{3}{*}{\textbf{v1}}}  & r $\leqslant$ 0.1       & 8.75                     & 8.97                     & 9.04                      & 9.06                     & \textbf{5.67}             \\ \cline{2-7} 
\multicolumn{1}{|l|}{}                              & r $\leqslant$ 0.2       & 16.22                    & 17.33                    & 17.90                     & 18.16                    & \textbf{9.25}            \\ \cline{2-7}
\multicolumn{1}{|l|}{}                              & r $\leqslant$ 0.4       & 15.15                     & 16.80                    & 17.88                     & 18.95                    & \textbf{9.75}            \\ \hline \hline
\multicolumn{1}{|c|}{\multirow{3}{*}{\textbf{v2}}}   & r $\leqslant$ 0.1       & 11.71                     & 11.49                     & 11.55                      & 11.57                     & \textbf{7.06}             \\ \cline{2-7}
\multicolumn{1}{|l|}{}                              & r $\leqslant$ 0.2       & 19.22                    & 17.76                    & 18.28                     & 18.56                    & \textbf{9.74}            \\ \cline{2-7}
\multicolumn{1}{|c|}{}                              & r $\leqslant$ 0.4       & 21.03                     & 17.19                    & 18.20                     & 19.44                    & \textbf{10.04}            \\ \hline \hline
\multicolumn{1}{|l|}{\multirow{3}{*}{\textbf{v3}}} & r $\leqslant$ 0.1       & 13.28                     & 14.23                     & 14.62                      & 14.78                     & \textbf{10.86}            \\ \cline{2-7}
\multicolumn{1}{|l|}{}                              & r $\leqslant$ 0.2       & 14.85                    & 17.27                    & 18.54                     & 19.49                    & \textbf{13.61}            \\ \cline{2-7}
\multicolumn{1}{|l|}{}                              & r $\leqslant$ 0.4       & \textbf{13.48}                     & 16.10                    & 17.72                     & 19.68                    & 14.73                     \\ \hline
\end{tabular}

}
\vspace{+3mm}
\caption{Neighborhood geodesic distance estimation MSE (x100) on the heldout ShapeNet training-category samples. We compare with \textit{KNN-Graph} based shortest path methods under different choices of $K$ values. \textit{Euc} represents the difference between Euclidean distance and geodesic distance. MSE(s) are reported under multiple radius ranges $r$. \textbf{v1} takes uniformly distributed point sets with 512 points as input, and \textbf{v2} uses randomly distributed point clouds. \textbf{v3} is tested using point clouds that have 2048 uniformly distributed points.}
\label{geo_est_train}
% \vspace{-3mm}
\end{table}
\begin{table}
\centering
\resizebox{0.95 \columnwidth}{!}{
% \makebox[\columnwidth][c]{

\begin{tabular}{lc|r|r|r|r|c|}
\cline{3-7}
                                                    & \multicolumn{1}{r|}{} & \multicolumn{1}{c|}{\textit{K-3}} & \multicolumn{1}{c|}{\textit{K-6}} & \multicolumn{1}{c|}{\textit{K-12}} & \multicolumn{1}{c|}{\textit{Euc}} & \multicolumn{1}{c|}{GeoNet} \\ \hline
\multicolumn{1}{|l|}{\multirow{2}{*}{\textbf{v1}}}  & r $\leqslant$ 0.1       & 8.81                     & 9.01                     & 9.05                      & 9.06                     & \textbf{7.52}             \\ \cline{2-7} 
\multicolumn{1}{|l|}{}                              & r $\leqslant$ 0.2       & 11.84                    & 12.88                    & 13.49                     & 13.75                    & \textbf{11.44}            \\ \hline \hline
\multicolumn{1}{|c|}{\multirow{2}{*}{\textbf{v2}}}   & r $\leqslant$ 0.1       & 10.52                    & 10.21                    & 10.25                     & 10.26                    & \textbf{8.94}             \\ \cline{2-7} 
\multicolumn{1}{|c|}{}                              & r $\leqslant$ 0.2       & 15.02                    & 12.99                    & 13.59                     & 13.86                    & \textbf{11.69}            \\ \hline \hline
\multicolumn{1}{|l|}{\multirow{2}{*}{\textbf{v3}}} & r $\leqslant$ 0.1       & 11.82                    & 12.39                    & 12.65                     & 12.75                    & \textbf{10.88}            \\ \cline{2-7} 
\multicolumn{1}{|l|}{}                              & r $\leqslant$ 0.2       & \textbf{11.80}           & 12.84                    & 13.55                     & 14.50                    & 12.26                     \\ \hline
\end{tabular}

}
\vspace{+3mm}
\caption{Geodesic neighborhood  estimation MSE (x100) on the leftout ShapeNet categories. \textbf{v1} takes uniformly distributed point sets with 512 points as input, and \textbf{v2} uses randomly distributed point clouds. \textbf{v3} is tested using point clouds that have 2048 uniformly distributed points.}
\label{geo_est_left_out}
% \vspace{+3mm}
\end{table}
%%%%%%%%%%%%%%%%%%%%%%%%%%%%%%%%%%%%%%%%%%%%%%%%%%%%%%%%%%%%%%%%%%%%%%%%%%%%%%%%%%%%%%

\textbf{PointNet++ Geodesic Fusion.} Fig.~\ref{geo_fusion_net} (bottom) illustrates the PointNet++ based fusion pipeline. Due to task as well as architecture differences between PU-Net and PointNet++, we make following changes to PUF to design a suitable fusion strategy that leverages PointNet++. First, for multi-scale grouping, we use the learned geodesic neighborhoods $\hat{G}_r(x_i)$ instead of Euclidean ones. Geodesic grouping brings attention to the underlying surfaces as well as structures of the point cloud. Second, while the PUF layer fuses estimated $\hat{G}_r(x_i) = \{\hat{g}_{ij}=\hat{d}_G(x_i,x_j) | x_j \in B_r(x_i)\}$, where $\hat{g}_{ij} \in {\mathbb R}$, of each neighborhood point set $B_r(x_i)$ into the backbone network, the POF layer uses the latent geodesic-aware features $\widetilde{\xi}_{ij} \in {\mathbb R^{\widetilde{C}}}$ extracted from the second-to-last FC layer in GeoNet. Namely, $\widetilde{\xi}_{ij}$ is an intermediate high-dimensional feature vector from $\xi_{ij}$ to $\hat{g}_{ij}$ via FC layers, and therefore it is better informed of the intrinsic point cloud topology. Third, in PointNet++ fusion we apply the POF layer in a hierarchical manner, leveraging farthest-point sampling. Thus, the learned features encode both local and global structural information of the point set. The total loss for POF also has two parts: One is for GeoNet training and the other is for the task-at-hand. We experiment on representative tasks that can benefit from understandings of the topological surface attributes. We use the $L_1$ error for point cloud normal estimation:

\begin{align} \label{loss_normal}
\begin{split}
L_{normal} = \sum_{x_i \in \chi} \sum_{j=1}^{3} \frac{\left | n_i^{(j)} - \hat{n}(x_i)^{(j)} \right |}{3N}
\end{split}
\end{align}

%%%%%%%%%%%%%%%%%%%%%%%%%%%%%%%%%%%%%%%%%%%%%%%%%%%%%%%%%%%%%%%%%%%%%%%%%%%%%%%%%%%%%%
\begin{figure*}
\vspace{-3mm}
\centering
\resizebox{0.85 \textwidth}{!}{
% \resizebox{1. \columnwidth}{!}{
% \makebox[1 \textwidth][c]{

% \includegraphics[height=6.5cm]{pr_curves.pdf}
\includegraphics[height=6.5cm]{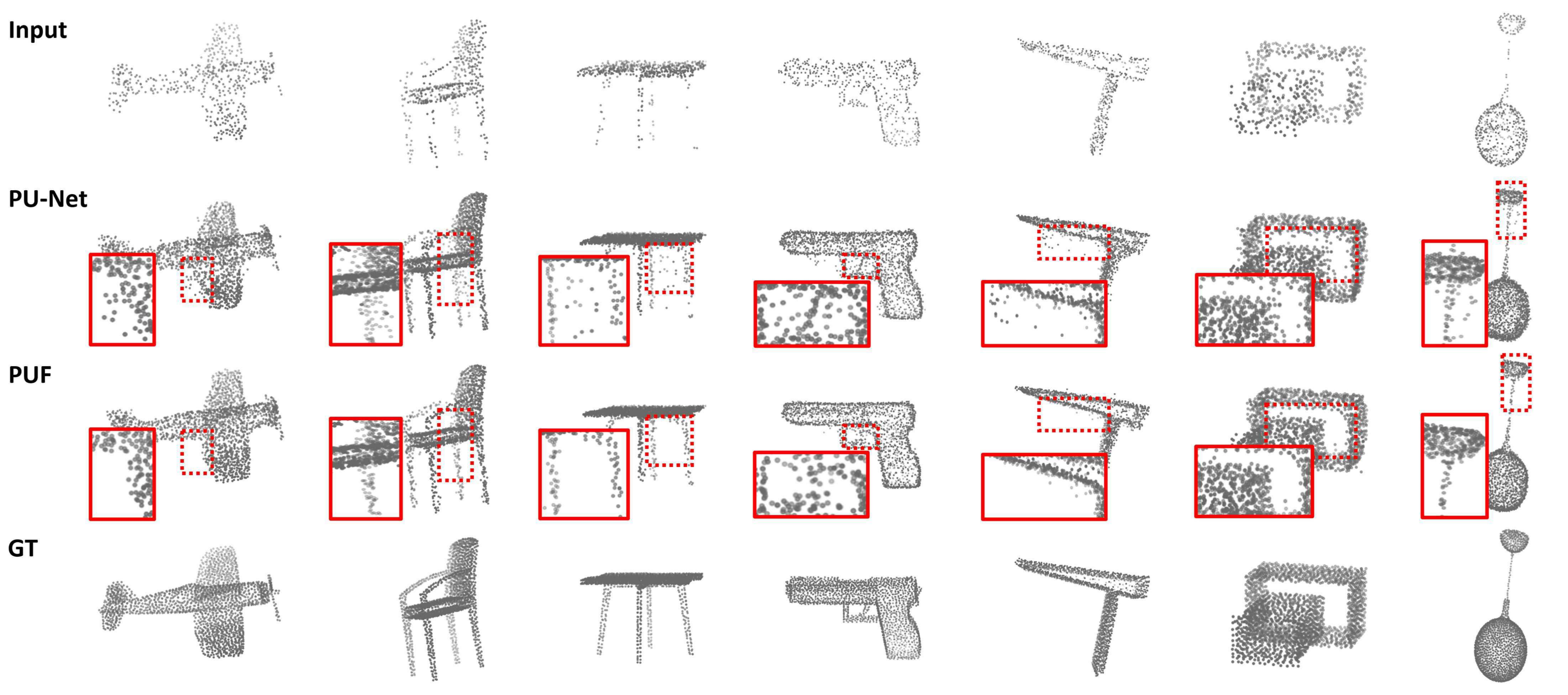}

}
\caption{Point cloud upsampling comparisons with PU-Net. The input point clouds have 512 points with random distributions and the upsampled point clouds have 2048 points. Red insets show details of the corresponding dashed region in the reconstruction.}
\label{upsampling_512to2048_shape_net}
% \vspace{-2mm}
\end{figure*}
%%%%%%%%%%%%%%%%%%%%%%%%%%%%%%%%%%%%%%%%%%%%%%%%%%%%%%%%%%%%%%%%%%%%%%%%%%%%%%%%%%%%%%

\noindent in which $n_i \in {\mathbb R^{3}}$ is the ground truth unit normal vector of $x_i$, and $\hat{n}(x_i) \in {\mathbb R^{3}}$ is the estimated normal. We then use the normal estimation to generate mesh via Poisson surface reconstruction~\cite{kazhdan2013screened}. To classify point clouds of non-rigid objects, we use cross-entropy loss:

\begin{align} \label{loss_nonrigid_cls}
\begin{split}
L_{cls} = -\sum_{c=1}^{S}y_{c}log(p_{c}(\chi))
\end{split}
\end{align}

\noindent where $S$ is the number of non-rigid object categories, and $c$ is class label; $y_c \in \{0,1\}$ is a binary indicator, which takes value $1$ if class label $c$ is ground truth for the input point set. $p_{c}(\chi) \in {\mathbb R}$ is the predicted probability w.r.t. class $c$ of the input point set.

\subsection{Implementation}
For GeoNet training, the multiscale loss $L_{geo_{l}}$ is enforced at three radius ranges: $0.1$, $0.2$ and $0.4$. We use Adam~\cite{kingma2014adam} with learning rate 0.001 and batchsize 3 for 8 epochs. To train the geodesic fusion networks, we set the task term weight $\lambda$ as 1, and use Adam with learning rate 0.0001 and batchsize 2 for around 300 to 1500 epochs depending on the task and the dataset. Source code in Tensorflow will be made available upon completion of the anonymous review process. % For point cloud upsampling, we train it for 400 epochs. For normal estimation and non-rigid shape classification we train for 400 and 1500 epochs, respectively.
%%%%%%%%% EXPERIMENTS
% %%%%%%%%%%%%%%%%%%%%%%%%%%%%%%%%%%%%%%%%%%%%%%%%%%%%%%%%%%%%%%%%%%%%%%%%%%%%%%%%%%%%%%
% \begin{table}
% \centering
% \resizebox{0.88 \columnwidth}{!}{
% % \makebox[\columnwidth][c]{

% \begin{tabular}{c|r|r|r|r|c|}
% \cline{2-6}
% \multicolumn{1}{r|}{}               & \multicolumn{1}{c|}{\textit{K-3}} & \multicolumn{1}{c|}{\textit{K-6}} & \multicolumn{1}{c|}{\textit{K-12}} & \multicolumn{1}{c|}{\textit{Euc}} & \multicolumn{1}{c|}{GeoNet} \\ \hline
% \multicolumn{1}{|c|}{r\ \leqslant 0.1} & 9.85                     & 10.50                     & 10.85                      & 11.00                     & \textbf{7.84}             \\ \hline
% \multicolumn{1}{|c|}{r\ \leqslant 0.2} & 10.19                    & 11.26                    & 12.03                     & 12.89                    & \textbf{8.74}             \\ \hline
% \multicolumn{1}{|c|}{r\ \leqslant 0.4} & \textbf{9.21}                     & 10.33                    & 11.33                     & 13.33                    & 9.34             \\ \hline
% \end{tabular}

% }
% \vspace{+3mm}
% \caption{Neighborhood geodesic distance estimation MSE (x100). The input point set has 2048 points that distributed uniformly.}
% \label{geo_est_2048_uniform}
% % \vspace{-4mm}
% \end{table}
% %%%%%%%%%%%%%%%%%%%%%%%%%%%%%%%%%%%%%%%%%%%%%%%%%%%%%%%%%%%%%%%%%%%%%%%%%%%%%%%%%%%%%%

\section{Experiments}

We put GeoNet to the test by estimating point cloud geodesic neighborhoods. To demonstrate the applicability of learned deep geodesic-aware representations, we also conduct experiments on down-stream point cloud tasks such as point upsampling, normal estimation, mesh reconstruction and non-rigid shape classification.

%%%%%%%%%%%%%%%%%%%%%%%%%%%%%%%%%%%%%%%%%%%%%%%%%%%%%%%%%%%%%%%%%%%%%%%%%%%%%%%%%%%%%%
\begin{figure}
\vspace{-5mm}
\centering
\resizebox{0.46 \textwidth}{!}{
% \makebox[1 \textwidth][c]{

% \includegraphics[height=6.5cm]{pr_curves.pdf}
\includegraphics[height=6.5cm]{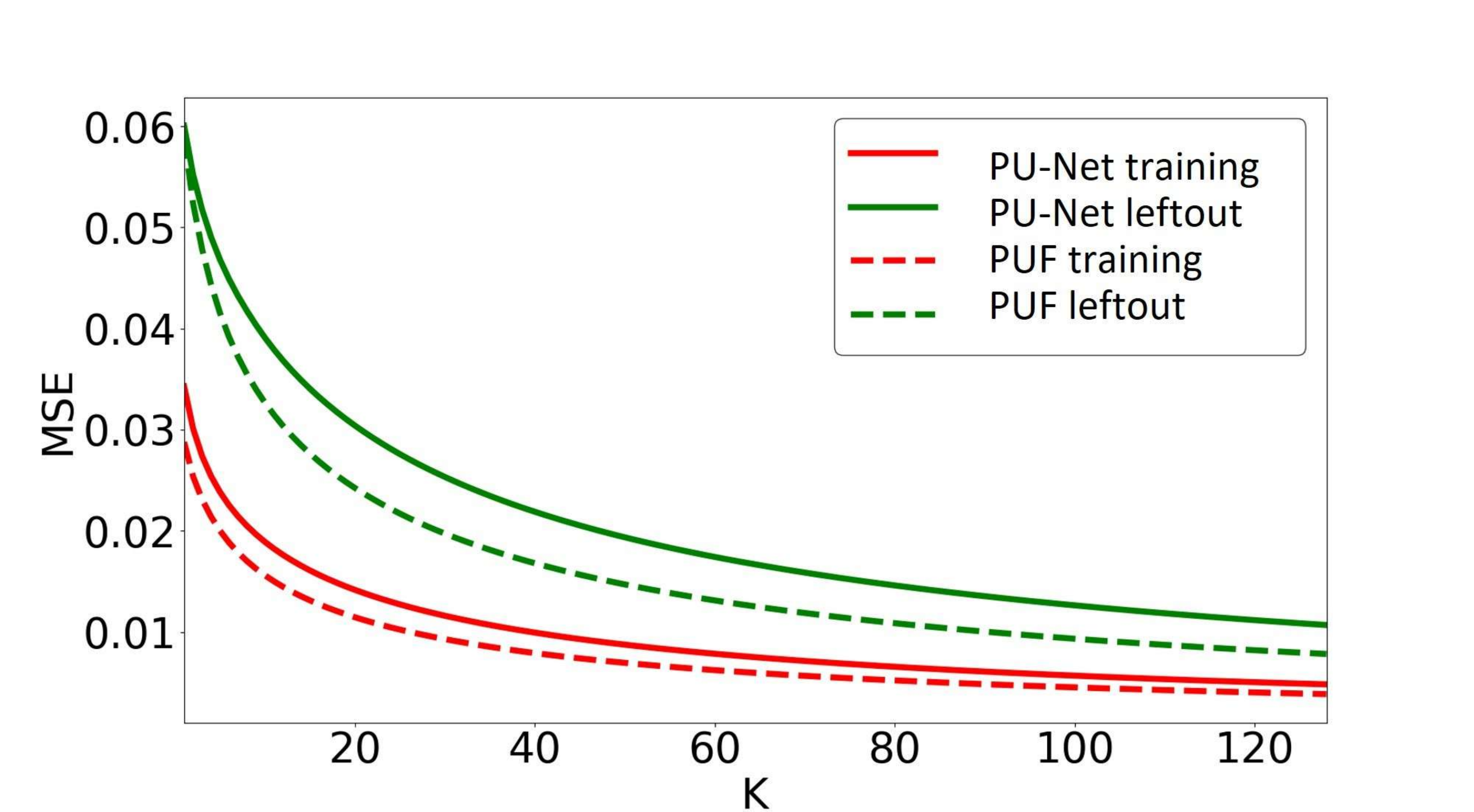}

}
\caption{Top-k mean square error (MSE) of upsampled points that have large errors, for both the heldout training-category samples (red) and the leftout ShapeNet categories (green).}
\label{top_256_mse_curves}
\end{figure}
%%%%%%%%%%%%%%%%%%%%%%%%%%%%%%%%%%%%%%%%%%%%%%%%%%%%%%%%%%%%%%%%%%%%%%%%%%%%%%%%%%%%%%

%%%%%%%%%%%%%%%%%%%%%%%%%%%%%%%%%%%%%%%%%%%%%%%%%%%%%%%%%%%%%%%%%%%%%%%%%%%%%%%%%%%%%%
\begin{table}
\centering
\resizebox{0.725 \columnwidth}{!}{
% \makebox[\columnwidth][c]{

\begin{tabular}{cc|c|c|c|}
\cline{3-5}
\multicolumn{1}{l}{}                            & \multicolumn{1}{l|}{} & MSE           & EMD           & CD            \\ \hline
\multicolumn{1}{|c|}{\multirow{2}{*}{Training}} & PU-Net                & 7.14          & 8.06          & 2.72          \\ \cline{2-5}
\multicolumn{1}{|c|}{}                          & PUF                  & \textbf{6.23} & \textbf{7.62} & \textbf{2.46} \\ \hline \hline
\multicolumn{1}{|c|}{\multirow{2}{*}{Leftout}}  & PU-Net                & 12.38         & 11.43         & 3.98          \\ \cline{2-5} 
\multicolumn{1}{|c|}{}                          & PUF                  & \textbf{9.55} & \textbf{8.90} & \textbf{3.27} \\ \hline
\end{tabular}

}
\vspace{+3mm}
\caption{Point cloud upsampling results on both the heldout training-category samples and the unseen ShapeNet categories. MSE(s) (x10000) are scaled for better visualization.}
\label{upsampling_values}
% \vspace{-1mm}
\end{table}
%%%%%%%%%%%%%%%%%%%%%%%%%%%%%%%%%%%%%%%%%%%%%%%%%%%%%%%%%%%%%%%%%%%%%%%%%%%%%%%%%%%%%%

%%%%%%%%%%%%%%%%%%%%%%%%%%%%%%%%%%%%%%%%%%%%%%%%%%%%%%%%%%%%%%%%%%%%%%%%%%%%%%%%%%%%%%
\begin{figure*}
\vspace{-2mm}
\centering
\resizebox{1.0 \textwidth}{!}{
% \makebox[1 \textwidth][c]{

% \includegraphics[height=6.5cm]{pr_curves.pdf}
\includegraphics[height=6.5cm]{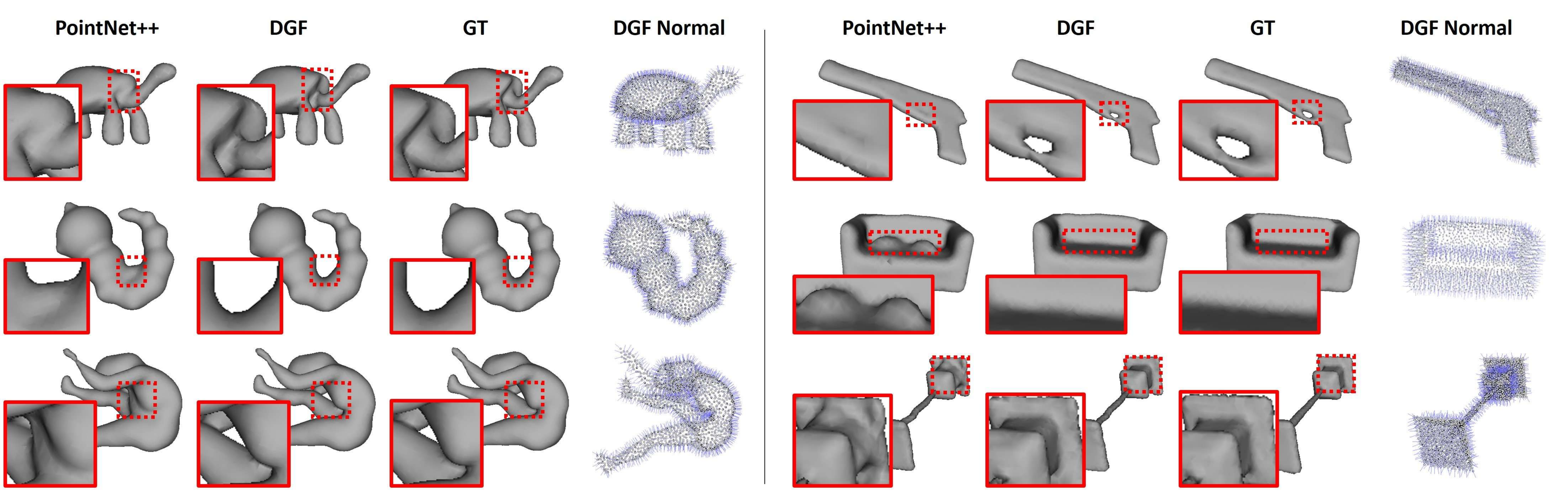}

}
% \vspace{+1mm}
\caption{Mesh reconstruction results on the Shrec15 (left) and the ShapeNet (right) datasets using the estimated normal by PointNet++ and our method POF. GT presents mesh reconstructed via the ground truth normal. We also visualize POF normal estimation in the fourth and the last columns.}
\label{mesh_shrec15_fullShapeNet}
% \vspace{-4mm}
\end{figure*}
%%%%%%%%%%%%%%%%%%%%%%%%%%%%%%%%%%%%%%%%%%%%%%%%%%%%%%%%%%%%%%%%%%%%%%%%%%%%%%%%%%%%%%

%%%%%%%%%%%%%%%%%%%%%%%%%%%%%%%%%%%%%%%%%%%%%%%%%%%%%%%%%%%%%%%%%%%%%%%%%%%%%%%%%%%%%%
\begin{figure}
\centering
\resizebox{0.86 \columnwidth}{!}{
% \makebox[1 \textwidth][c]{

% \includegraphics[height=6.5cm]{pr_curves.pdf}
\includegraphics[height=6.5cm]{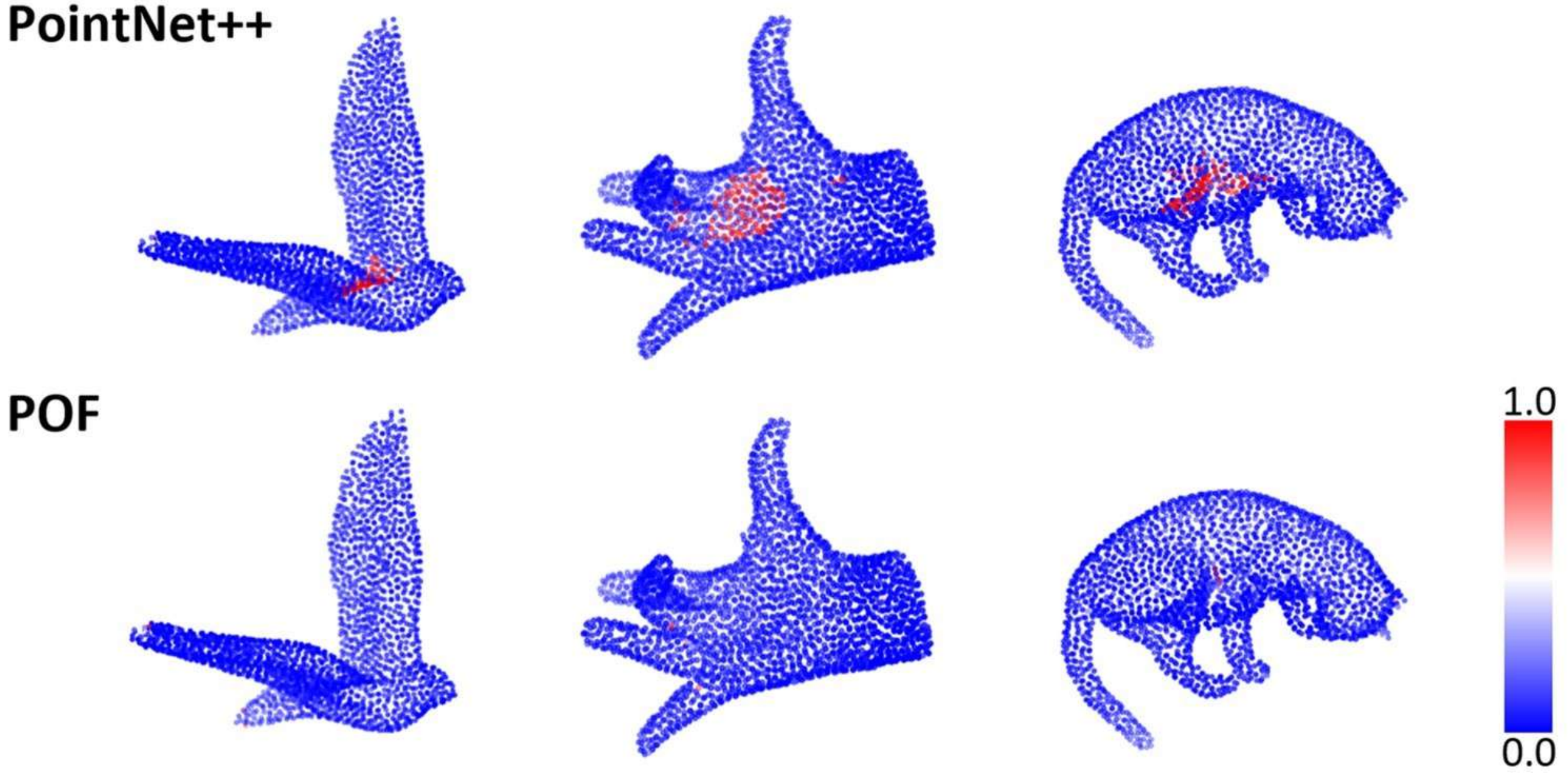}

}
\caption{Point set normal estimation errors. Blue indicates small errors and red is for large ones.}
\label{normal_est_shrec15}
% \vspace{-4mm}
\end{figure}
%%%%%%%%%%%%%%%%%%%%%%%%%%%%%%%%%%%%%%%%%%%%%%%%%%%%%%%%%%%%%%%%%%%%%%%%%%%%%%%%%%%%%%

\subsection{Geodesic Neighborhood Estimation} In Tab.~\ref{geo_est_train} (\textbf{v1}) we show geodesic distance set, $G_r(x_i)$, estimation results on the ShapeNet dataset~\cite{chang2015shapenet} using point clouds with 512 uniformly distributed points. Mean-squared errors (MSE) are reported under multiple radius scales $r$ w.r.t. $x_i \in \chi$. GeoNet demonstrates consistent improvement over the baselines. Representative results are visualized in Fig.~\ref{geo_est_shape_net}. Our method captures various topological patterns, such as curved surfaces, layered structures, inner/outer parts, etc.

\textbf{Generality.} We test GeoNet's robustness under different point set distributions and sizes. In Tab.~\ref{geo_est_train} (\textbf{v2}) we use point clouds with 512 randomly distributed points as input. We also test on dense point sets that contain 2048 uniformly distributed points in Tab.~\ref{geo_est_train} (\textbf{v3}). Our results are robust to different point set distributions as well as sizes. To show the generalization performance, in Tab.~\ref{geo_est_left_out} we report results on the leftout ShapeNet categories. Our method performs better on unseen categories, while \textit{KNN-Graph} based shortest path approaches suffer from point set distribution randomness, density changes and unsuitable choices of \textit{K} values.

%%%%%%%%%%%%%%%%%%%%%%%%%%%%%%%%%%%%%%%%%%%%%%%%%%%%%%%%%%%%%%%%%%%%%%%%%%%%%%%%%%%%%%
\begin{table}
\centering
\resizebox{1 \columnwidth}{0.038 \textheight}{
% \makebox[\columnwidth][c]{

\begin{tabular}{c|r|r|r|r|}
\cline{2-5}
                                 & \multicolumn{1}{c|}{$\leqslant$ 2.5$^{\circ}$} & \multicolumn{1}{c|}{$\leqslant$ 5$^{\circ}$} & \multicolumn{1}{c|}{$\leqslant$ 10$^{\circ}$} & \multicolumn{1}{c|}{$\leqslant$ 15$^{\circ}$} \\ \hline
\multicolumn{1}{|c|}{PCA}        & 6.16$\pm$0.01               & 14.85$\pm$0.02            & 27.16$\pm$0.17             & 34.17$\pm$0.28             \\ \hline
\multicolumn{1}{|c|}{PointNet++} & 12.81$\pm$0.18              & 33.37$\pm$0.92            & 61.58$\pm$2.02             & 75.49$\pm$1.95             \\ \hline
\multicolumn{1}{|c|}{POF}       & \textbf{16.26}$\pm$0.30              & \textbf{39.02}$\pm$1.09            & \textbf{66.98}$\pm$1.46             & \textbf{79.66}$\pm$1.21             \\ \hline
\end{tabular}

}
\vspace{+1mm}
\caption{Point cloud normal estimation accuracy (\%) on the Shrec15 dataset under multiple angle thresholds.}
\label{shrec_normal_est}
% \vspace{-4mm}
\end{table}
%%%%%%%%%%%%%%%%%%%%%%%%%%%%%%%%%%%%%%%%%%%%%%%%%%%%%%%%%%%%%%%%%%%%%%%%%%%%%%%%%%%%%%

%%%%%%%%%%%%%%%%%%%%%%%%%%%%%%%%%%%%%%%%%%%%%%%%%%%%%%%%%%%%%%%%%%%%%%%%%%%%%%%%%%%%%%
\begin{table}
\centering
\resizebox{0.95 \columnwidth}{!}{
% \makebox[\columnwidth][c]{

\begin{tabular}{cc|r|r|r|r|}
\cline{3-6}
                                                &            & $\leqslant$ 2.5$^{\circ}$            & $\leqslant$ 5$^{\circ}$              & $\leqslant$ 10$^{\circ}$             & $\leqslant$ 15$^{\circ}$             \\ \hline
\multicolumn{1}{|c|}{\multirow{3}{*}{Training}} & PCA        & 5.33           & 10.11          & 18.52          & 24.82          \\ \cline{2-6} 
\multicolumn{1}{|l|}{}                          & PointNet++ & 30.68          & 43.19          & 55.91          & 62.30          \\ \cline{2-6} 
\multicolumn{1}{|l|}{}                          & POF       & \textbf{32.04} & \textbf{45.02} & \textbf{57.52} & \textbf{63.62} \\ \hline \hline
\multicolumn{1}{|c|}{\multirow{3}{*}{Leftout}}  & PCA        & 5.24           & 10.59          & 18.99          & 25.17          \\ \cline{2-6} 
\multicolumn{1}{|l|}{}                          & PointNet++ & 17.35          & 28.82          & 43.26          & 51.17          \\ \cline{2-6} 
\multicolumn{1}{|l|}{}                          & POF       & \textbf{19.13} & \textbf{31.83} & \textbf{46.22} & \textbf{53.78} \\ \hline
\end{tabular}

}
\vspace{+3mm}
\caption{Point cloud normal estimation accuracy (\%) on the ShapeNet dataset for both heldout training-category samples and leftout categories.}
\label{shape_net_normal_est}
% \vspace{-4mm}
\end{table}
%%%%%%%%%%%%%%%%%%%%%%%%%%%%%%%%%%%%%%%%%%%%%%%%%%%%%%%%%%%%%%%%%%%%%%%%%%%%%%%%%%%%%%

\subsection{Point Cloud Upsampling} We test PUF on point cloud upsampling and present results in Tab.~\ref{upsampling_values}. We compare against the state-of-the-art point set upsampling method PU-Net on three metrics: MSE, EMD as well as the Chamfer Distance (CD). Our method outperforms the baseline under all metrics by 9.25\% average improvement on the heldout training-category samples. Since geodesic neighborhoods are better informed of the underlying point set topology than Euclidean ones, PUF upsampling produces less outliers and recovers more details in Fig.~\ref{upsampling_512to2048_shape_net}, such as curves and sharp structures.

\textbf{Generality.} To analyze outlier robustness (i.e. points with large reconstruction errors), we plot top-k MSE in Fig.~\ref{top_256_mse_curves}. Our method generates fewer outliers on both the heldout training-category samples and the unseen categories. We also report quantitative results on the leftout categories in Tab.~\ref{upsampling_values}. Again, PUF significantly surpasses the state-of-the-art upsampling method PU-Net under three different evaluation metrics.

\subsection{Normal Estimation and Mesh Reconstruction}
For normal estimation we apply PointNet++ geodesic fusion, POF, then we conduct Poisson mesh reconstruction leveraging the estimated normals. Quantitative results for normal estimation on the Shrec15 dataset and the ShapeNet dataset are given in Tab.~\ref{shrec_normal_est} and Tab.~\ref{shape_net_normal_est}, respectively. We compare our method with the traditional PCA algorithm as well as the state-of-the-art deep learning method PointNet++. Our results outperform the baselines by around 10\% relative improvement. In Fig.~\ref{normal_est_shrec15}, we visualize typical normal estimation errors, showing that PointNet++ usually fails at high-curvature and complex-surface regions. For further evidence, we visualize Poisson mesh reconstruction in Fig.~\ref{mesh_shrec15_fullShapeNet} using the estimated normals.

\begin{figure}
% \vspace{+2.5mm}
\centering
\resizebox{0.88 \columnwidth}{!}{
% \makebox[1 \textwidth][c]{

% \includegraphics[height=6.5cm]{pr_curves.pdf}
\includegraphics[height=6.5cm]{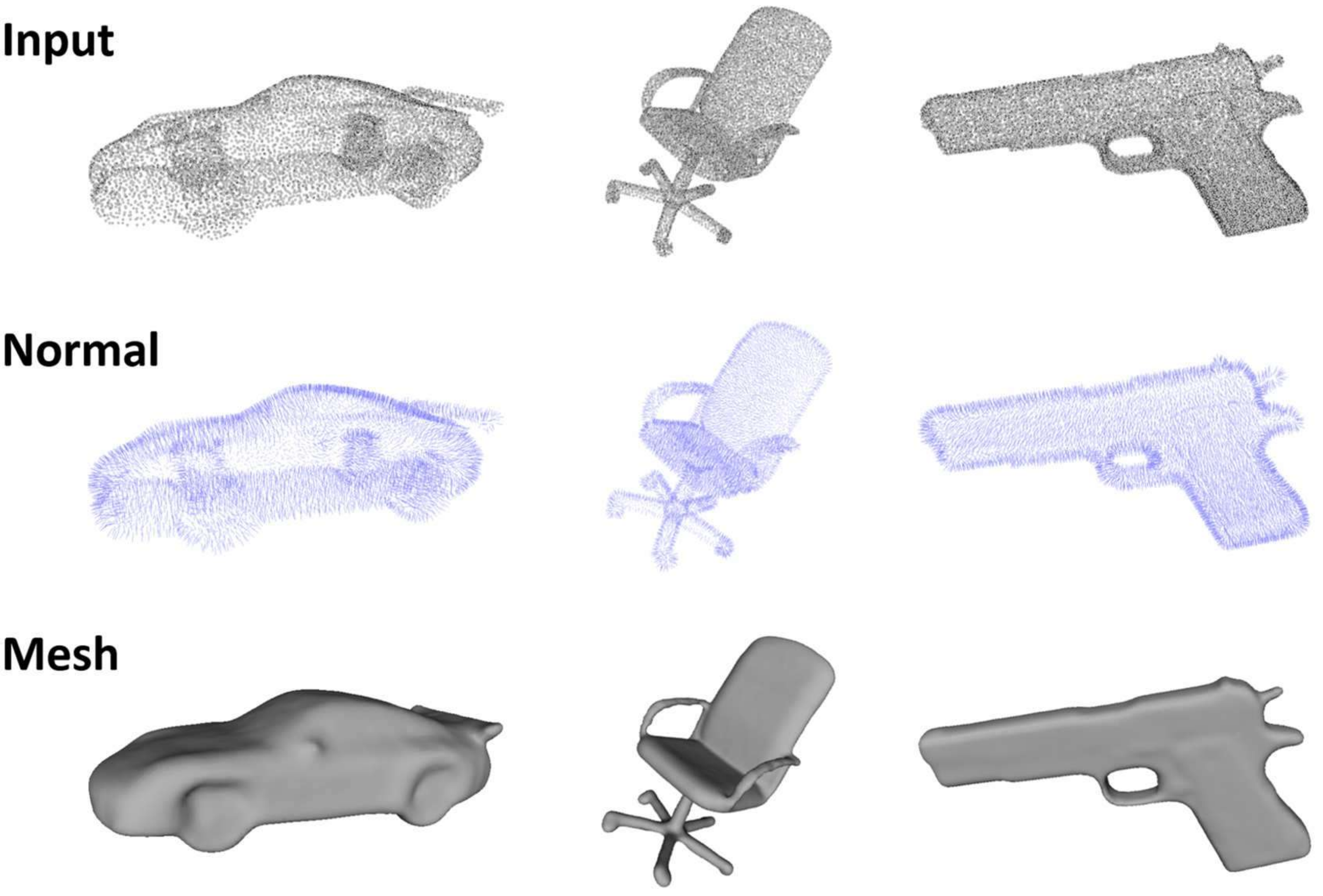}

}
\caption{Normal estimation and Poisson mesh reconstruction results by POF using dense point clouds with 8192 points.}
\label{mesh_fullShapeNetx4}
% \vspace{-3mm}
\end{figure}

\textbf{Generality.} In Tab.~\ref{shape_net_normal_est} we evaluate normal estimation performance on the leftout ShapeNet categories. Our method has higher accuracy over competing methods under multiple angle thresholds. Though trained with point clouds of 2048 points, POF is also tested on denser input. In Fig.~\ref{mesh_fullShapeNetx4} we take point clouds with 8192 points as input, and visualize the normal estimation and mesh reconstruction results, which shows that our method generalizes to dense point clouds without re-training and produces fine-scaled mesh. 

% %%%%%%%%%%%%%%%%%%%%%%%%%%%%%%%%%%%%%%%%%%%%%%%%%%%%%%%%%%%%%%%%%%%%%%%%%%%%%%%%%%%%%%
% \begin{figure*}
% \centering
% \resizebox{1.0 \textwidth}{!}{
% % \makebox[1 \textwidth][c]{

% % \includegraphics[height=6.5cm]{pr_curves.pdf}
% \includegraphics[height=6.5cm]{fig/mesh_shrec15.png}

% }
% \caption{Mesh reconstruction on the Shrec15 dataset using the estimated normal.}
% \label{mesh_shrec15}
% \end{figure*}
% %%%%%%%%%%%%%%%%%%%%%%%%%%%%%%%%%%%%%%%%%%%%%%%%%%%%%%%%%%%%%%%%%%%%%%%%%%%%%%%%%%%%%%

% %%%%%%%%%%%%%%%%%%%%%%%%%%%%%%%%%%%%%%%%%%%%%%%%%%%%%%%%%%%%%%%%%%%%%%%%%%%%%%%%%%%%%%
% \begin{figure*}
% \centering
% \resizebox{1.0 \textwidth}{!}{
% % \makebox[1 \textwidth][c]{

% % \includegraphics[height=6.5cm]{pr_curves.pdf}
% \includegraphics[height=6.5cm]{fig/mesh_fullShapeNet.png}

% }
% \caption{Mesh reconstruction on the shapeNet dataset using the estimated normal.}
% \label{mesh_fullShapeNet}
% \end{figure*}
% %%%%%%%%%%%%%%%%%%%%%%%%%%%%%%%%%%%%%%%%%%%%%%%%%%%%%%%%%%%%%%%%%%%%%%%%%%%%%%%%%%%%%%

%%%%%%%%%%%%%%%%%%%%%%%%%%%%%%%%%%%%%%%%%%%%%%%%%%%%%%%%%%%%%%%%%%%%%%%%%%%%%%%%%%%%%%
%%%%%%%%%%%%%%%%%%%%%%%%%%%%%%%%%%%%%%%%%%%%%%%%%%%%%%%%%%%%%%%%%%%%%%%%%%%%%%%%%%%%%%

%%%%%%%%%%%%%%%%%%%%%%%%%%%%%%%%%%%%%%%%%%%%%%%%%%%%%%%%%%%%%%%%%%%%%%%%%%%%%%%%%%%%%%
\begin{table}
% \vspace{+1mm}
\centering
\resizebox{0.7 \columnwidth}{!}{
% \makebox[\columnwidth][c]{

\begin{tabular}{c|c|c|}
\cline{2-3}
\multicolumn{1}{l|}{}            & Input feature      & Accuracy (\%)  \\ \hline
\multicolumn{1}{|c|}{PointNet++} & XYZ                & 73.56          \\ \hline
\multicolumn{1}{|c|}{POF}       & XYZ                & \textbf{94.67} \\ \hline \hline
\multicolumn{1}{|c|}{DeepGM}     & Intrinsic features & 93.03          \\ \hline
\end{tabular}

}
\vspace{+3mm}
\caption{Point cloud classification of non-rigid shapes on the Shrec15 dataset.}
\label{shrec15_cls}
% \vspace{-1mm}
\end{table}
%%%%%%%%%%%%%%%%%%%%%%%%%%%%%%%%%%%%%%%%%%%%%%%%%%%%%%%%%%%%%%%%%%%%%%%%%%%%%%%%%%%%%%

%%%%%%%%%%%%%%%%%%%%%%%%%%%%%%%%%%%%%%%%%%%%%%%%%%%%%%%%%%%%%%%%%%%%%%%%%%%%%%%%%%%%%%
\begin{table}
% \vspace{+1mm}
\centering
\resizebox{0.812 \columnwidth}{!}{
% \makebox[\columnwidth][c]{

\begin{tabular}{c|c|c|c|c|c|}
\cline{2-6}
\multicolumn{1}{l|}{}            & \multicolumn{5}{c|}{Gaussian Noise Level}   \\ \cline{2-6} 
                                 & 0.8\% & 0.9\% & 1.0\% & 1.1\% & 1.2\% \\ \hline
\multicolumn{1}{|c|}{PointNet++} & 70.54 & 69.27 & 67.83 & 65.66 & 62.38 \\ \hline
\multicolumn{1}{|c|}{POF}        & 91.89 & 90.93 & 89.40  & 87.72 & 84.98 \\ \hline
\end{tabular}

}
\vspace{+3mm}
\caption{Noisy point clouds classification accuracy (\%). We add Gaussian
noise of 0.8\% to 1.2\% of unit ball radius.}
\label{noisy_cls}
% \vspace{-4mm}
\end{table}
%%%%%%%%%%%%%%%%%%%%%%%%%%%%%%%%%%%%%%%%%%%%%%%%%%%%%%%%%%%%%%%%%%%%%%%%%%%%%%%%%%%%%%

\subsection{Non-rigid Shape Classification}

Results of non-rigid shape classification are reported in Tab.~\ref{shrec15_cls}. While POF and PointNet++ only take point cloud-based $xyz$ Euclidean coordinates as input, DeepGM requires offline computed intrinsic features from mesh data in the ground truth geodesic metric space. Though using less informative data, our method has higher classification accuracy than other methods, which further demonstrates that the proposed geodesic fusion architecture, POF, is suitable for solving tasks that require understandings of the underlying point cloud surface attributes.

\textbf{Generality.} We add Gaussian noise of different levels to the input and conduct noisy point clouds classification. Comparisons are shown in Tab.~\ref{noisy_cls}. POF outperforms PointNet++ under several noise levels. Our method also demonstrates better noise robustness. It shows a 10.24\% decrease in relative accuracy at the maximum noise level, while PointNet++ decreases by up to 15.20\%.

\subsection{Failure Modes}
\begin{figure}
\vspace{-2mm}
\centering
\resizebox{0.9 \columnwidth}{!}{
% \makebox[1 \textwidth][c]{

% \includegraphics[height=6.5cm]{pr_curves.pdf}
\includegraphics[height=6.5cm]{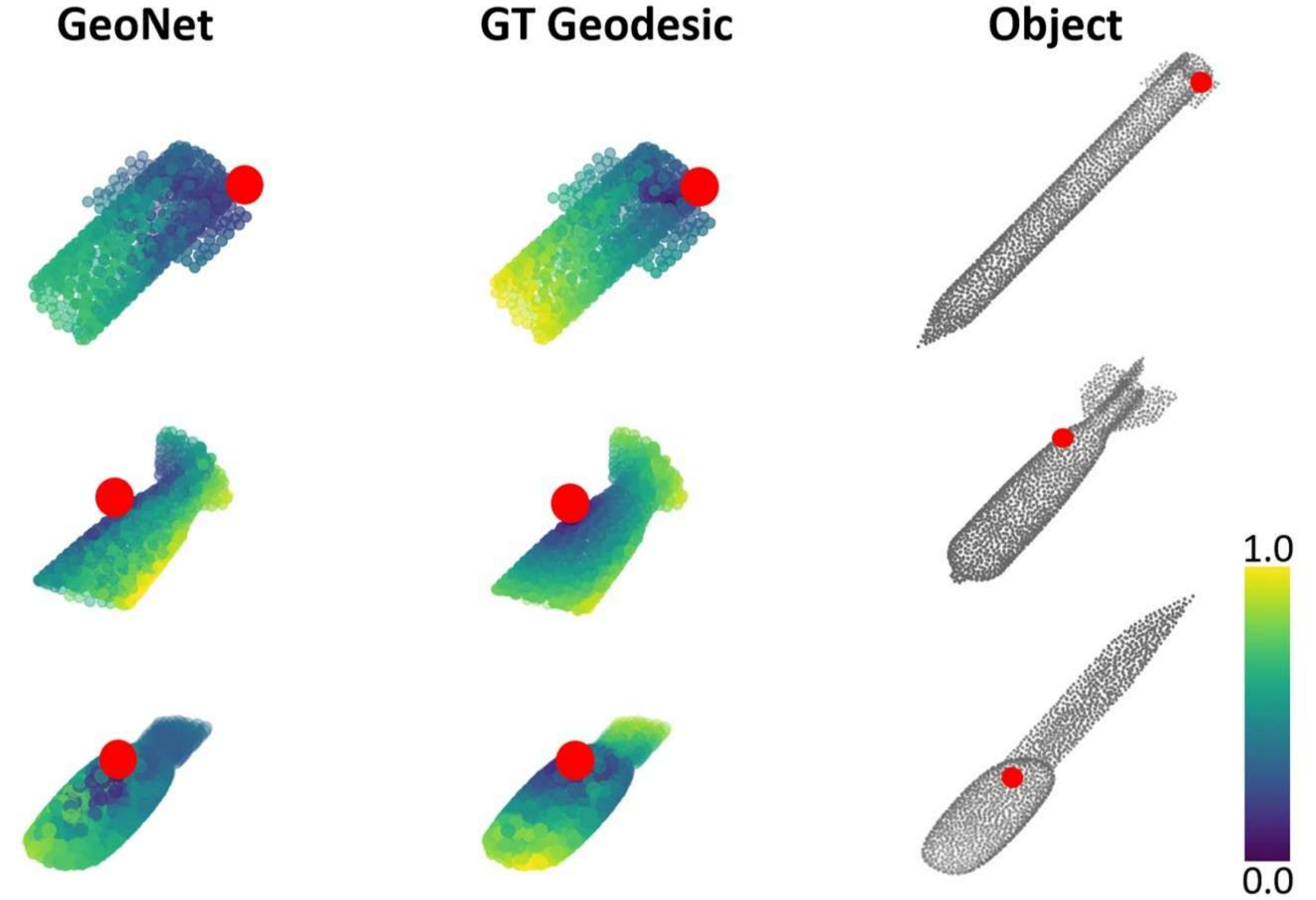}

}
\caption{Failure cases of geodesic neighborhood estimation for stick-shaped objects (e.g. rocket, knife, etc.) which have large ratios between length and width/height. Red dots indicate the reference point. Points in dark-purple are closer to the reference point than those in bright-yellow.}
\label{failure}
% \vspace{-4mm}
\end{figure}

Failure cases of geodesic neighborhood estimation are shown in Fig.~\ref{failure}. Due to large ratios between length and width/height, after normalizing a stick-shaped object (e.g. rocket, knife, etc.) into a unit ball we need high precision small values to represent its point-pair geodesic distance along the width/height sides. Since stick-shaped objects like rocket and knife only take up a small portion of the training data, GeoNet tends to make mistakes for heldout samples from these categories at inference time. We have not found additional failure cases, and quantitative improvements continue to take effect due to rich surface-based topological information learned during the geodesic-supervised training process.

%%%%%%%%%%%%%%%%%%%%%%%%%%%%%%%%%%%%%%%%%%%%%%%%%%%%%%%%%%%%%%%%%%%%%%%%%%%%%%%%%%%%%%
%%%%%%%%%%%%%%%%%%%%%%%%%%%%%%%%%%%%%%%%%%%%%%%%%%%%%%%%%%%%%%%%%%%%%%%%%%%%%%%%%%%%%%
%%%%%%%%% CONCLUSION
\section{Conclusion}

We have presented GeoNet, a novel deep learning architecture to learn the geodesic space-based topological structure within point clouds. The training process is supervised by the ground truth geodesic distance and therefore the learned representations reflect the intrinsic structure of the underlying point set surfaces. To demonstrate the applicability of such a topology estimation network, we also propose fusion methods to incorporate GeoNet into computational schemes that involve the standard backbone architectures for point cloud analysis. Our method is tested on both geometric and semantic tasks and outperforms the state-of-the-art methods, including point upsampling, normal estimation, mesh reconstruction and non-rigid shape classification.

{\small
\bibliographystyle{ieee}
\bibliography{egbib}

\begin{thebibliography}{10}\itemsep=-1pt

\bibitem{alexa2003computing}
M.~Alexa, J.~Behr, D.~Cohen-Or, S.~Fleishman, D.~Levin, and C.~T. Silva.
\newblock Computing and rendering point set surfaces.
\newblock {\em IEEE Transactions on visualization and computer graphics},
  9(1):3--15, 2003.

\bibitem{amenta1999surface}
N.~Amenta and M.~Bern.
\newblock Surface reconstruction by voronoi filtering.
\newblock {\em Discrete \& Computational Geometry}, 22(4):481--504, 1999.

\bibitem{aubry2011wave}
M.~Aubry, U.~Schlickewei, and D.~Cremers.
\newblock The wave kernel signature: A quantum mechanical approach to shape
  analysis.
\newblock In {\em Computer Vision Workshops (ICCV Workshops), 2011 IEEE
  International Conference on}, pages 1626--1633. IEEE, 2011.

\bibitem{cazals2005estimating}
F.~Cazals and M.~Pouget.
\newblock Estimating differential quantities using polynomial fitting of
  osculating jets.
\newblock {\em Computer Aided Geometric Design}, 22(2):121--146, 2005.

\bibitem{chang2015shapenet}
A.~X. Chang, T.~Funkhouser, L.~Guibas, P.~Hanrahan, Q.~Huang, Z.~Li,
  S.~Savarese, M.~Savva, S.~Song, H.~Su, et~al.
\newblock Shapenet: An information-rich 3d model repository.
\newblock {\em arXiv preprint arXiv:1512.03012}, 2015.

\bibitem{chen1990shortest}
J.~Chen and Y.~Han.
\newblock Shortest paths on a polyhedron.
\newblock In {\em Proceedings of the sixth annual symposium on Computational
  geometry}, pages 360--369. ACM, 1990.

\bibitem{crane2013geodesics}
K.~Crane, C.~Weischedel, and M.~Wardetzky.
\newblock Geodesics in heat: A new approach to computing distance based on heat
  flow.
\newblock {\em ACM Transactions on Graphics (TOG)}, 32(5):152, 2013.

\bibitem{dai2017shape}
A.~Dai, C.~R. Qi, and M.~Nie{\ss}ner.
\newblock Shape completion using 3d-encoder-predictor cnns and shape synthesis.
\newblock In {\em Proc. IEEE Conf. on Computer Vision and Pattern Recognition
  (CVPR)}, volume~3, 2017.

\bibitem{defferrard2016convolutional}
M.~Defferrard, X.~Bresson, and P.~Vandergheynst.
\newblock Convolutional neural networks on graphs with fast localized spectral
  filtering.
\newblock In {\em Advances in Neural Information Processing Systems}, pages
  3844--3852, 2016.

\bibitem{Dijkstra}
E.~W. Dijkstra.
\newblock A note on two problems in connexion with graphs.
\newblock {\em Numer. Math.}, 1(1):269--271, Dec. 1959.

\bibitem{floyd1962algorithm}
R.~W. Floyd.
\newblock Algorithm 97: shortest path.
\newblock {\em Communications of the ACM}, 5(6):345, 1962.

\bibitem{garland1997surface}
M.~Garland and P.~S. Heckbert.
\newblock Surface simplification using quadric error metrics.
\newblock In {\em Proceedings of the 24th annual conference on Computer
  graphics and interactive techniques}, pages 209--216. ACM
  Press/Addison-Wesley Publishing Co., 1997.

\bibitem{guennebaud2007algebraic}
G.~Guennebaud and M.~Gross.
\newblock Algebraic point set surfaces.
\newblock In {\em ACM Transactions on Graphics (TOG)}, volume~26, page~23. ACM,
  2007.

\bibitem{guerrero2018pcpnet}
P.~Guerrero, Y.~Kleiman, M.~Ovsjanikov, and N.~J. Mitra.
\newblock Pcpnet learning local shape properties from raw point clouds.
\newblock In {\em Computer Graphics Forum}, volume~37, pages 75--85. Wiley
  Online Library, 2018.

\bibitem{han2015matchnet}
X.~Han, T.~Leung, Y.~Jia, R.~Sukthankar, and A.~C. Berg.
\newblock Matchnet: Unifying feature and metric learning for patch-based
  matching.
\newblock In {\em Proceedings of the IEEE Conference on Computer Vision and
  Pattern Recognition}, pages 3279--3286, 2015.

\bibitem{han2017high}
X.~Han, Z.~Li, H.~Huang, E.~Kalogerakis, and Y.~Yu.
\newblock High-resolution shape completion using deep neural networks for
  global structure and local geometry inference.
\newblock In {\em Proceedings of IEEE International Conference on Computer
  Vision (ICCV)}, 2017.

\bibitem{hoppe1992surface}
H.~Hoppe, T.~DeRose, T.~Duchamp, J.~McDonald, and W.~Stuetzle.
\newblock {\em Surface reconstruction from unorganized points}, volume~26.
\newblock ACM, 1992.

\bibitem{huang2009consolidation}
H.~Huang, D.~Li, H.~Zhang, U.~Ascher, and D.~Cohen-Or.
\newblock Consolidation of unorganized point clouds for surface reconstruction.
\newblock {\em ACM transactions on graphics (TOG)}, 28(5):176, 2009.

\bibitem{johnson1977efficient}
D.~B. Johnson.
\newblock Efficient algorithms for shortest paths in sparse networks.
\newblock {\em Journal of the ACM (JACM)}, 24(1):1--13, 1977.

\bibitem{jolliffe2011principal}
I.~Jolliffe.
\newblock Principal component analysis.
\newblock In {\em International encyclopedia of statistical science}, pages
  1094--1096. Springer, 2011.

\bibitem{kazhdan2013screened}
M.~Kazhdan and H.~Hoppe.
\newblock Screened poisson surface reconstruction.
\newblock {\em ACM Transactions on Graphics (ToG)}, 32(3):29, 2013.

\bibitem{kingma2014adam}
D.~P. Kingma and J.~Ba.
\newblock Adam: A method for stochastic optimization.
\newblock {\em arXiv preprint arXiv:1412.6980}, 2014.

\bibitem{lipman2007parameterization}
Y.~Lipman, D.~Cohen-Or, D.~Levin, and H.~Tal-Ezer.
\newblock Parameterization-free projection for geometry reconstruction.
\newblock {\em ACM Transactions on Graphics (TOG)}, 26(3):22, 2007.

\bibitem{luciano2018deep}
L.~Luciano and A.~B. Hamza.
\newblock Deep learning with geodesic moments for 3d shape classification.
\newblock {\em Pattern Recognition Letters}, 105:182--190, 2018.

\bibitem{masoumi2016spectral}
M.~Masoumi, C.~Li, and A.~B. Hamza.
\newblock A spectral graph wavelet approach for nonrigid 3d shape retrieval.
\newblock {\em Pattern Recognition Letters}, 83:339--348, 2016.

\bibitem{mitchell1987discrete}
J.~S. Mitchell, D.~M. Mount, and C.~H. Papadimitriou.
\newblock The discrete geodesic problem.
\newblock {\em SIAM Journal on Computing}, 16(4):647--668, 1987.

\bibitem{qi2017pointnet}
C.~R. Qi, H.~Su, K.~Mo, and L.~J. Guibas.
\newblock Pointnet: Deep learning on point sets for 3d classification and
  segmentation.
\newblock {\em Proc. Computer Vision and Pattern Recognition (CVPR), IEEE},
  1(2):4, 2017.

\bibitem{qi2017pointnet++}
C.~R. Qi, L.~Yi, H.~Su, and L.~J. Guibas.
\newblock Pointnet++: Deep hierarchical feature learning on point sets in a
  metric space.
\newblock In {\em Advances in Neural Information Processing Systems}, pages
  5099--5108, 2017.

\bibitem{reuter2006laplace}
M.~Reuter, F.-E. Wolter, and N.~Peinecke.
\newblock Laplace--beltrami spectra as ‘shape-dna’of surfaces and solids.
\newblock {\em Computer-Aided Design}, 38(4):342--366, 2006.

\bibitem{sharir1986shortest}
M.~Sharir and A.~Schorr.
\newblock On shortest paths in polyhedral spaces.
\newblock {\em SIAM Journal on Computing}, 15(1):193--215, 1986.

\bibitem{sun2009concise}
J.~Sun, M.~Ovsjanikov, and L.~Guibas.
\newblock A concise and provably informative multi-scale signature based on
  heat diffusion.
\newblock In {\em Computer graphics forum}, volume~28, pages 1383--1392. Wiley
  Online Library, 2009.

\bibitem{surazhsky2005fast}
V.~Surazhsky, T.~Surazhsky, D.~Kirsanov, S.~J. Gortler, and H.~Hoppe.
\newblock Fast exact and approximate geodesics on meshes.
\newblock In {\em ACM transactions on graphics (TOG)}, volume~24, pages
  553--560. Acm, 2005.

\bibitem{wu2016learning}
J.~Wu, C.~Zhang, T.~Xue, B.~Freeman, and J.~Tenenbaum.
\newblock Learning a probabilistic latent space of object shapes via 3d
  generative-adversarial modeling.
\newblock In {\em Advances in Neural Information Processing Systems}, pages
  82--90, 2016.

\bibitem{wu20153d}
Z.~Wu, S.~Song, A.~Khosla, F.~Yu, L.~Zhang, X.~Tang, and J.~Xiao.
\newblock 3d shapenets: A deep representation for volumetric shapes.
\newblock In {\em Proceedings of the IEEE conference on computer vision and
  pattern recognition}, pages 1912--1920, 2015.

\bibitem{xin2009improving}
S.-Q. Xin and G.-J. Wang.
\newblock Improving chen and han's algorithm on the discrete geodesic problem.
\newblock {\em ACM Transactions on Graphics (TOG)}, 28(4):104, 2009.

\bibitem{yi2017syncspeccnn}
L.~Yi, H.~Su, X.~Guo, and L.~J. Guibas.
\newblock Syncspeccnn: Synchronized spectral cnn for 3d shape segmentation.
\newblock In {\em CVPR}, pages 6584--6592, 2017.

\bibitem{yu2018pu}
L.~Yu, X.~Li, C.-W. Fu, D.~Cohen-Or, and P.-A. Heng.
\newblock Pu-net: Point cloud upsampling network.
\newblock In {\em Proceedings of the IEEE Conference on Computer Vision and
  Pattern Recognition}, pages 2790--2799, 2018.

\end{thebibliography}
}

\end{document}